\documentclass[a4paper,conference]{IEEEtran}
\IEEEoverridecommandlockouts
\usepackage{amsmath,amssymb,amsfonts}
\usepackage{algorithmic}
\usepackage{booktabs}
\usepackage{caption}
\usepackage{capt-of}
\usepackage{comment}
\usepackage{color}
\usepackage{graphicx}
\usepackage[pdftex]{hyperref}
\usepackage{multirow}
\usepackage{nameref}
\usepackage{placeins}
\usepackage{subfigure}
\usepackage[table,xcdraw]{xcolor}
\usepackage{tabularx}
\usepackage{textcomp}
\usepackage[noadjust]{cite}
\def\BibTeX{{\rm B\kern-.05em{\sc i\kern-.025em b}\kern-.08em
		T\kern-.1667em\lower.7ex\hbox{E}\kern-.125emX}}

\begin{document}
	
	\title{Label Noise Types and Their Effects on Deep Learning}
	
	\author{
		\IEEEauthorblockN{G\"{o}rkem Algan\IEEEauthorrefmark{1}\IEEEauthorrefmark{2}, Ilkay Ulusoy\IEEEauthorrefmark{1}}
		\IEEEauthorblockA{\IEEEauthorrefmark{1}Middle East Technical University, Electrical-Electronics Engineering
			\\\{e162565, ilkay\}@metu.edu.tr}
		\IEEEauthorblockA{\IEEEauthorrefmark{2}ASELSAN
			\\galgan@aselsan.com.tr}
	}
	
	\maketitle
	
	\begin{abstract}
        The recent success of deep learning is mostly due to the availability of big datasets with clean annotations. However, gathering a cleanly annotated dataset is not always feasible due to practical challenges. As a result, label noise is a common problem in datasets, and numerous methods to train deep neural networks in the presence of noisy labels are proposed in the literature. These methods commonly use benchmark datasets with synthetic label noise on the training set. However, there are multiple types of label noise, and each of them has its own characteristic impact on learning. Since each work generates a different kind of label noise, it is problematic to test and compare those algorithms in the literature fairly. In this work, we provide a detailed analysis of the effects of different kinds of label noise on learning. Moreover, we propose a generic framework to generate feature-dependent label noise, which we show to be the most challenging case for learning. Our proposed method aims to emphasize similarities among data instances by sparsely distributing them in the feature domain. By this approach, samples that are more likely to be mislabeled are detected from their softmax probabilities, and their labels are flipped to the corresponding class. The proposed method can be applied to any clean dataset to synthesize feature-dependent noisy labels. For the ease of other researchers to test their algorithms with noisy labels, we share corrupted labels for the most commonly used benchmark datasets. Our code and generated noisy synthetic labels are available online\footnote{Github repository: \url{https://github.com/spidy390/corrupting_labels_with_distillation}}.
	\end{abstract}
	
	
	\section{Introduction} \label{introduction}
Recent advancement in deep learning has led to great improvements in computer vision systems \cite{krizhevsky2012imagenet,he2016deep,simonyan2014very}. Even though it is shown that deep networks have an impressive ability to generalize \cite{rolnick2017deep,flatow2017robustness,drory2018neural}, these powerful models have a great tendency to memorize even complete random noise \cite{zhang2016understanding,krueger2017deep,arpit2017closer}. Therefore, avoiding memorization is an important challenge to be overcame in order to obtain representative neural networks and it gets even more crucial in the presence of noise \cite{drory2018neural,hataya2018investigating}. There are two types of noise, namely: feature noise and label noise \cite{frenay2014classification}. Generally speaking, label noise is considered to be more harmful than feature noise \cite{zhu2004class}.

It is known that label noise has a negative impact on the training process for broad range of applications \cite{zhu2004class, nettleton2010study, pechenizkiy2006class,flatow2017robustness}. As a result, various methods are proposed in the literature to prevent performance degradation caused by noisy labels \cite{algan2019image,frenay2014classification}. However, it is problematic to test those algorithms when the dataset is noisy and there is no ground truth available. Therefore, commonly adopted methodology is to add synthetic label noise to training set of an available benchmark dataset while keeping test set clean. Since there is no generic framework to corrupt labels of the given dataset in a systematic way to mimic real-world noise, each work adds their own characteristic label noise. This results in subjective evaluation of algorithms and prevents fair comparison of the methods. Even though literature generally considers noisy labels phenomenon as one compact problem, there are multiple types of label noise \cite{frenay2014classification} (\textit{uniform noise, class-dependent noise, feature-dependent noise}) and each has its own characteristic affect on performance. 

Negative effect of label noise depending on its type is an understudied problem. Works of \cite{flatow2017robustness} focuses on uniform label noise and its impact on learning solely. However, as shown on \autoref{experiments}, uniform label noise can easily be handled by neural networks without an extra modification. Therefore, it can be misleading to draw conclusion about label noise by experimenting only with uniform label noise. In their unpublished work \cite{hataya2018investigating}, authors consider the case of class-dependent and class-independent noise. But, this is still and uncompleted picture, since feature dependent noise is not considered. Also, it is shown in \autoref{experiments} that feature-dependent noise is more harmful than class-dependent noise. To the best of our knowledge, there is no work devoted to fully investigate the negative effects of all types of label noise on deep networks. In this work, each of these label noise types are analyzed and their effects on learning process is investigated. We show that, each one has its own characteristic effect on learning and therefore should be evaluated under their own category.

Even though some benchmark datasets with noisy training set and clean test set exists \cite{rolnick2017deep}, literature still commonly adopts the approach off corrupting labels synthetically. The main reason for this approach is the ability to generate corrupted datasets from toy datasets in purpose of quick application and testing of proposed algorithm. Yet, question of how to add synthetic noise in order to test noise robust algorithms stays to be an open question. For this purpose, most human-like noise type is feature-dependent noise, in which features of each instance effects probability of being mislabeled. For example, in cars dataset, some sport cars are more similar to classic cars than others. These specific instances have greater chance to be mislabeled by a human annotator. However, this relation among specific instances is not considered in uniform or class-dependent noise. This work proposes a feature-dependent label corruption algorithm, that is inspired by \textit{knowled-distillation} technique \cite{hinton2015distilling}. Our methodology aims to learn representations which would result in sparse distribution of data instances in feature domain. At the end, similarities among data samples are extracted and labels are flipped for uncertain samples. Additionally, we provide pre-generated noisy labels for commonly used datasets (MNIST\cite{lecun1998mnist}, Fashion-MNIST \cite{xiao2017fashion}, CIFAR10 \cite{torralba200880}, CIFAR100 \cite{krizhevsky2009learning}) in purpose of other researchers to test their label noise robust algorithms. 

Our contribution to literature can be summarized as follows
\begin{itemize}
    \item Detailed analysis of negative impacts of different types of label noise on learning.
    \item Label corruption algorithm that utilizes data similarities in feature domain by creating sparse representation of data.
\end{itemize}

This paper is organized as follows. In \autoref{preliminary}, different label noise types are explained. Proposed solution to generate synthetic noise is given in \autoref{generatingnoise}. Section \ref{experiments} discusses the effects of different types of label noise on learning. Finally, \autoref{conclusion} concludes the paper.	
	\section{Preliminary} \label{preliminary}
This chapter presents label noise types and their implementations in literature. Label noise types can be subdivided into three main groups as follows.

\begin{itemize}
    \item \textbf{\textit{Uniform noise:}} Flipping probability of label from its true class to any other class is equally distributed. Many works in literature use synthetic uniform label noise by just flipping labels randomly for a given percentage of data instances \cite{ma2018dimensionality,jenni2018deep,jindal2016learning,yuan2018iterative}. 
    \item \textbf{\textit{Class-dependent noise:}} Flipping probability of label depends on the true class of the data instance. This is mostly represented by a confusion matrix and can be designed in different ways. The easiest way is to attain inter-class transition probabilities just random \cite{litany2018soseleto}, so that there is still class dependence since transition probabilities are given according to classes but without any correlation to class similarities. In a more structured way, noise transition matrix can be designed in a way that similar classes have a bigger probability to be flipped to each other \cite{patrini2017making,junnan2018learning,yi2019probabilistic,sukhbaatar2014training,hendrycks2018using}. Some works use pairwise noise, in which transition from one class can only be defined to one another class \cite{han2018co,ren2018learning,goldberger2016training,han2018pumpout,yu2019disagreement}. Work of \cite{xu2019l_dmi} checks the popularity of classes and constructs transition matrix so that mislabeling happens from popular class to unpopular class or vice versa. 
    \item \textbf{\textit{Feature-dependent noise:}} The probability of mislabeling depends on features of instances. In order to generate feature-dependent noise, features of each instance should be extracted, and their similarities to other instances from different classes should be evaluated. Unlike uniform and class-dependent noise, there are much fewer implementations of synthetic feature-dependent label noise. One particular work in this field is \cite{inouye2017hyperparameter}, where data is clustered with the kNN algorithm, and labels are flipped randomly for clusters of data. This method provides concentrated noise in the feature space. But, this type of synthetic noise doesn't evaluate the instance similarities and therefore different from our proposed approach. Alternatively, in case there is a surrounding text for each image in the dataset, some works create noisy labels from the interpretations of these texts \cite{khetan2017learning,vahdat2017toward,li2017learning,misra2016seeing}, assuming surrounding texts are related to features of data. But this approach is restricted to datasets with surrounding user-defined texts, which is not the case for most of the time.
\end{itemize}

In this work, we focus on the closed set problem, in which all data instances are from the given class set. However, some works also investigated open set problem \cite{wang2018iterative}, where the dataset is polluted with instances that don't belong to any class from the class set.
	\section{Generating Synthetic Noise} \label{generatingnoise}
This section presents the methodologies to produce different types of synthetic label noise. \textit{Uniform noise} and \textit{class-dependent noise} can be represented with noise transition matrix $N$ where $N_{ij}$ represents the probability of flipping label from class $i$ to $j$. Since noise transition matrix consists of probabilities, $\sum_{j}N_{ij}=1$. On the other hand, in \textit{feature-dependent noise}, each instance has its own transition probability depending on its features. Therefore, it can not be generated using a noise transition matrix. The following sections will describe the process of generating these types of noises. Generated noisy labels are visualized with T-SNE plots in \autoref{fig:tsneplots}.

\begin{figure*}[h]
  \centering
  \includegraphics[width=\textwidth]{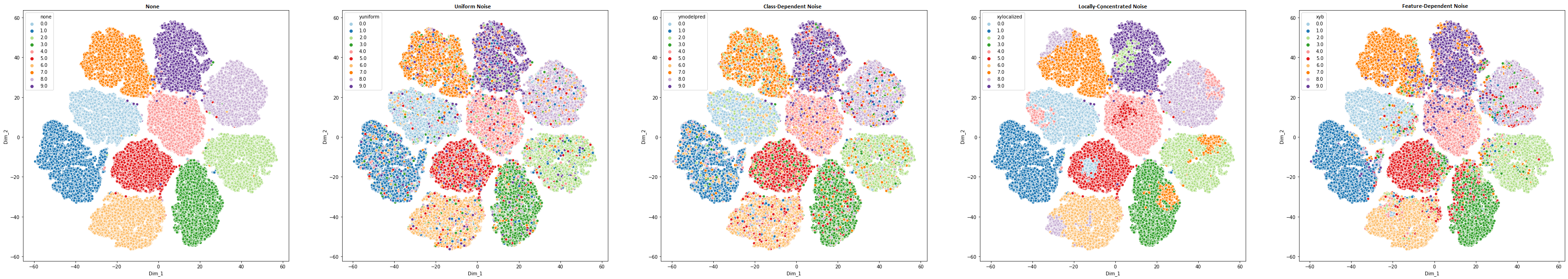}
  \caption{T-SNE plot of data distribution in feature space for 25\% noise ratio. a) Cleanly annotated data b) Uniform label noise c) Class-dependent label noise d) Locally concentrated label noise \cite{inouye2017hyperparameter} e) Feature-dependent label noise generated with distillation. As can be observed, in feature dependent noise, mostly, samples that are close to decision boundaries are corrupted.}
\end{figure*} \label{fig:tsneplots}

\subsection{Uniform Noise}
For this type of noise, each entry in the noise transition matrix, besides diagonal ones, are equally distributed. Noise transition matrix can be defined as follows:

\begin{equation}
    N_{ij} =
      \begin{cases}
        p & \text{if } i = j\\
        \dfrac{1-p}{M-1} & \text{if } i != j
      \end{cases}       
\end{equation}
where $M$ is the number of classes.

\subsection{Class-dependent Noise}
As mentioned earlier, there are various types of class-dependent noise. In this work, we consider structured class-dependent noise, where similar classes have a higher probability of being mislabeled with each other. In order to generate a structured confusion matrix, a deep network is trained on the training set, and its confusion matrix is calculated depending on its predictions on the test set. This confusion matrix is then used for the noise transition matrix.

\subsection{Feature-dependent noise}
Compared to the previous two label noise models, feature dependent noise is harder to implement since all samples should be vectorized in the feature domain, and similarities among samples should be calculated. \cite{inouye2017hyperparameter} flips labels for clusters of instances, which results in locally concentrated label noise. However, this approach doesn't utilize the similarities among instances; therefore, different from our proposed method. 

One option is to train a deep network on the dataset first and then use it as the feature extractor. However, since the network extracts the features of data that it is trained on, it is prone to overfitting. Since we are especially interested in similarities among instances in feature space, it is desired that samples are sparsely distributed. On the contrary, in the case of overfitting, samples are gathered in a small region in feature space.

Therefore, in this work we used the idea of \textit{knowledge distillation} \cite{hinton2015distilling}. In the original work, the authors used distillation to transfer knowledge from the big teacher network to a much smaller student network without decreasing the performance. The idea is mainly motivated by learning from soft labels where the similarity of each instance to each class is emphasized by \textit{temperature} hyperparameter. 

Class probabilities on softmax output, beyond the true class probability, are usually very low. But, compared with each other, some classes may have a much higher probability than others, and this carries important information about that data instance, which is also called as \textit{dark knowledge}. By making probability distribution smoother, this relation is emphasized, as shown in \autoref{eq:temp}.

\begin{equation}
    q_i=\dfrac{exp(z_i/T)}{\sum_{j}exp(z_j/T)} \label{eq:temp}
\end{equation}

Instead of being trained on hard labels, the student network is trained on the weighted sum of hard labels and soft labels produced by the teacher network. So, the loss function is defined as follows,

\begin{equation}
    L(y_i,f(x_i))=\alpha l(y_i,f(x_i)) + (1-\alpha)l(q_i,f(x_i)) \label{eq:distilloss}
\end{equation}

where $q_i$ represents the soft labels produced by the teacher network using temperature $T$ and $y_i$ represents the given label.

Within the context of this work, we are not interested in compressing the network to a smaller network. However, the idea of learning by emphasizing instance similarities can be used to find instances that have similar features with other classes. For that purpose, our student network is at the same size as the teacher network. Firstly, the teacher network is trained on the dataset. Secondly, soft labels are produced from the softmax output of the teacher for a given temperature $T$. Thirdly, the student network is trained on soft labels and hard labels with a weighting factor $\alpha$ in the loss function. Finally, by checking softmax probabilities of instances, samples that have similar features to other classes are detected, and their labels are flipped to corresponding classes.

To see if the proposed method results in a more sparse distribution of data in feature space, we can check variance of instances belonging to classes and average over each class as follows

\begin{equation}
  \sigma=\dfrac{\sum_{i}^{N}var(Q_i)}{M}\label{eq:var}
\end{equation}

where $M$ is the number of classes and $Q_i$ is the feature matrix of instances belonging to class $i$. Features are extracted from the layer output before the softmax layer.

For straightforward training on the MNIST dataset, the network manages to get 99\% test accuracy while having $\sigma=78.3$. On the other hand, the network trained with distillation achieves 95\% accuracy while having $\sigma=563.5$. Its accuracy is comparable to the original network, but learned representations are much more sparse, which is useful to extract similarities.
	\section{Experiments} \label{experiments}
We performed our experiments on different datasets under various conditions to inspect the effect of label noise under different setups. When simulating label noise, training-set and validation-set are corrupted with the same type of noise, while keeping the test set clean. With this setup, the validation set is used to measure the transferability of learned representations from noisy training set to the noisy validation set, since it means high accuracy on the validation set. On the other hand, the test set is used to evaluate the true performance of the network. The following subsections present the results of various experiments. All experiments are run on Intel i7, GTX 1080, 32GB RAM computer. Our code is available at \url{https://github.com/spidy390/corrupting_labels_with_distillation}. 

\subsection{Datasets and Models}
\textbf{MNIST-Fashion:} We have used CNN with two convolutional layers followed by a max-pooling layer and two fully-connected layers. There are dropout layers after the max-pooling layer and the first fully connected layer. Categorical cross-entropy is used as the loss function and stochastic gradient descent with learning rate 0.1 and momentum 0.9 as the optimizer. The learning rate is halved at every 20 epochs.

\textbf{CIFAR100:} We have used a model based on VGG16 architecture. Categorical cross-entropy is used as the loss function and stochastic gradient descent with learning rate 0.1 and momentum 0.9 as the optimizer. The learning rate is halved at every 20 epochs.

\subsection{Noise Types}
\autoref{fig:noisetypes} shows accuracies for various noise types. In this setup training set and validation set are corrupted with the same noise while keeping the test set clean. As can be seen from the figures, both proposed feature-dependent noise and locally-concentrated noise causes most performance degradation on the test set. We observed that the network fits locally-concentrated noise slowest. This is expected behavior since it is harder to find decision boundaries due to corrupted clusters of data. Interestingly, proposed feature-dependent noise gives almost the same behavior as no noise case, in terms of convergence speed and validation accuracy. However, it results in much worse performance on the test set. Therefore, it can be concluded that, in the case of feature-dependent noise and absence of a clean test set, it is harder to evaluate whether the network is overfitting the noisy samples or not.

\begin{figure}[h]
  \centering
  \includegraphics[width=\columnwidth]{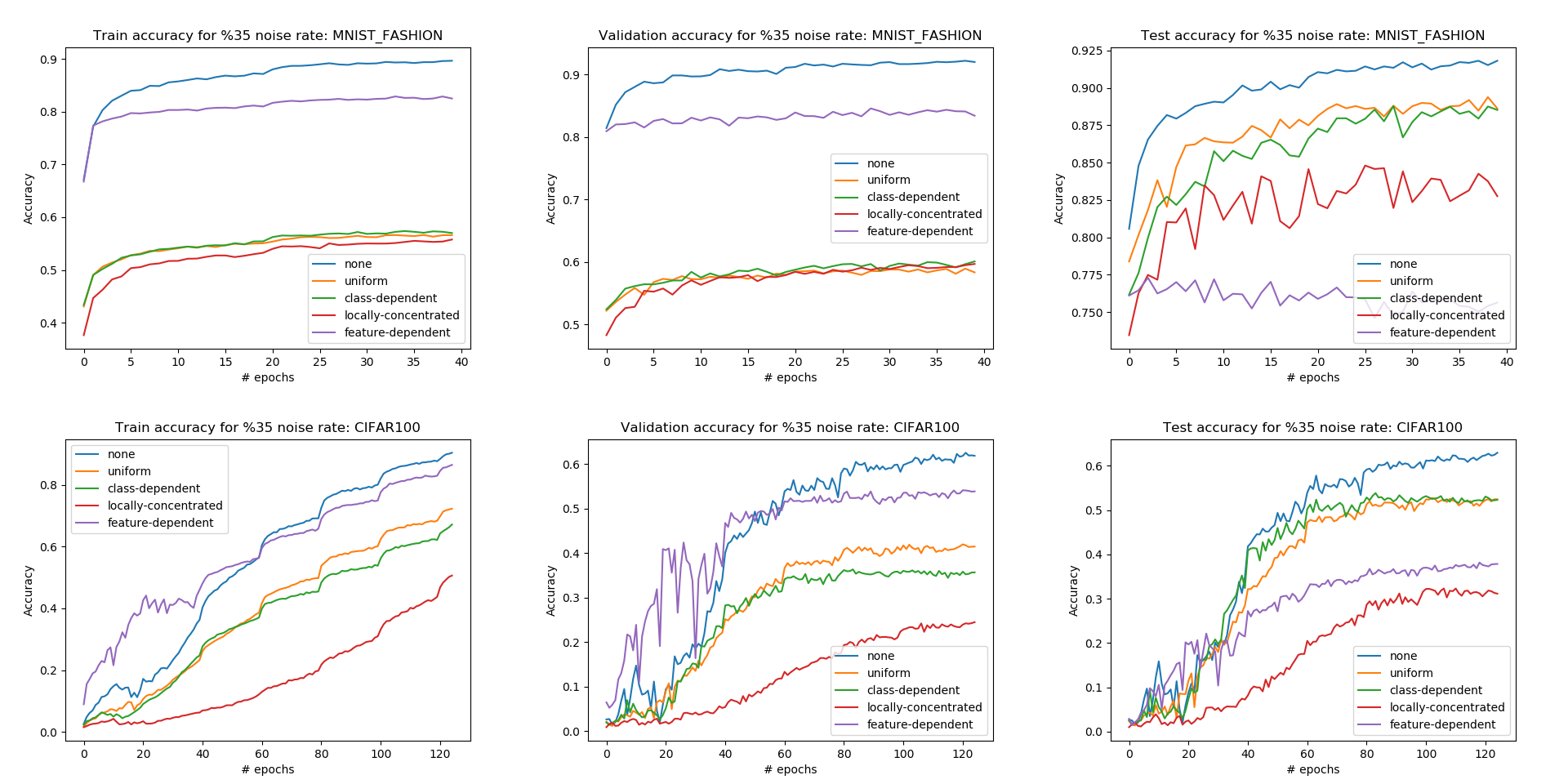}
  \caption{Train, validation and test accuracies for different noise types. Noise types tested are uniform noise, class-dependent noise, locally concentrated noise and feature-dependent noise. The upper row is train, validation and test accuracies for MNIST-Fashion at noise rate 35\%. The lower row is train, validation and test accuracies for CIFAR100 at noise rate 35\%.}
\end{figure} \label{fig:noisetypes}

\begin{table}[]
  \centering
  \begin{tabular}{|l|cccc|}
  \hline
  \textbf{Data}(\%) &
    \textbf{Uniform} &
    \textbf{\begin{tabular}[c]{@{}c@{}}Class\\ Dependent\end{tabular}} &
    \textbf{\begin{tabular}[c]{@{}c@{}}Locally\\ Concentrated\end{tabular}} &
    \multicolumn{1}{l|}{\textbf{\begin{tabular}[c]{@{}l@{}}Feature\\ Dependent\end{tabular}}} \\ \hline
  \textbf{1}   & 70\% & 62\% & 65\% & 66\% \\ \hline
  \textbf{10}  & 82\% & 76\% & 74\% & 77\% \\ \hline
  \textbf{40}  & 88\% & 86\% & 81\% & 75\% \\ \hline
  \textbf{70}  & 89\% & 87\% & 83\% & 75\% \\ \hline
  \textbf{100} & 89\% & 89\% & 83\% & 76\% \\ \hline
  \end{tabular}
  \caption{Test accuracies for MNIST-Fashion for various training dataset sizes for noise ratio of 35\%.}
\end{table} \label{tbl:datasize}

\subsection{Noise Ratios}
For each type of noise, we corrupted labels of the dataset for various amounts, from 5\% to 75\% with steps of 10\%. Results on MNIST-Fashion in \autoref{tbl:noiseratios}, show that feature-dependent noise gives the highest accuracy for train and validation sets for all noise ratios. On the other hand, in almost all noise ratios, feature-dependent noise results in the worst performance on the clean test set. Differently, on a more complex CIFAR100 dataset, it is seen in \autoref{fig:noiseratios} that locally-concentrated noise results in the worst performance. Surprisingly, for feature-dependent noise, in all noise ratios, we observed that the network can achieve around 90\% accuracy on the training set. This can be explained as follows; since noise is added near decision boundaries, the network can easily fit the noisy data by expanding clusters. Moreover, validation accuracy doesn't decrease with increasing noise ratio as well. This indicates that learned representations are consistent with the feature-dependent noise itself.

\begin{figure}[h]
	\centering
	\includegraphics[width=\columnwidth]{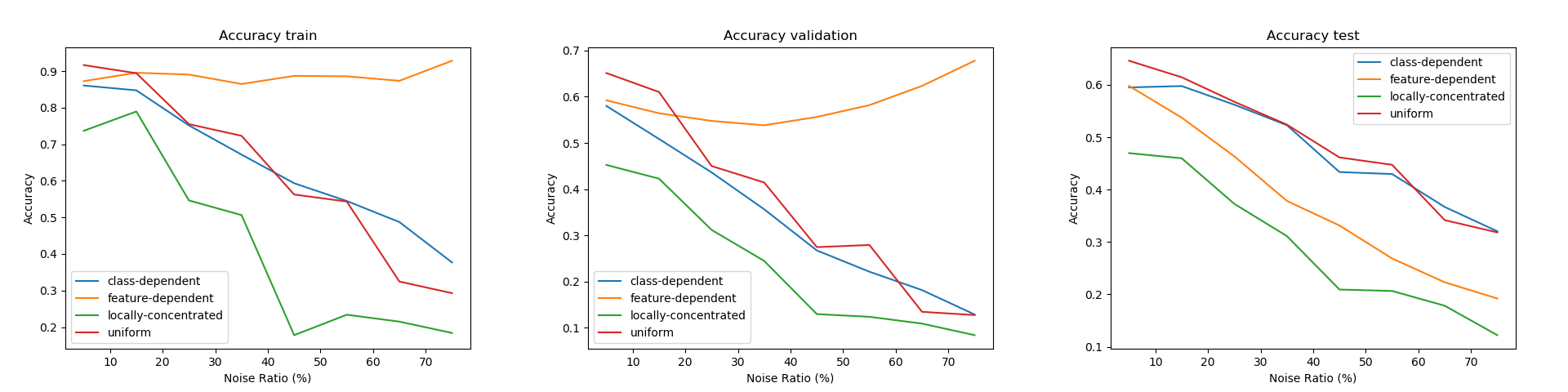}
	\caption{Train, validation and test accuracies for CIFAR100 at different noise ratios.}
\end{figure} \label{fig:noiseratios}

\begin{table*}[]
  \centering
  \begin{tabular}{|l|llll|llll|llll|}
  \hline
   &
    \multicolumn{4}{c}{\textbf{Train}} &
    \multicolumn{4}{c|}{\textbf{Validation}} &
    \multicolumn{4}{c|}{\textbf{Test}} \\ \hline
  \textbf{\begin{tabular}[c]{@{}l@{}}Noise\\  Ratio \\ (\%)\end{tabular}} &
    \textbf{Uniform} &
    \textbf{\begin{tabular}[c]{@{}l@{}}Class \\ Dep.\end{tabular}} &
    \textbf{\begin{tabular}[c]{@{}l@{}}Locally \\ Con.\end{tabular}} &
    \textbf{\begin{tabular}[c]{@{}l@{}}Feature \\ Dep.\end{tabular}} &
    \textbf{Uniform} &
    \textbf{\begin{tabular}[c]{@{}l@{}}Class\\ Dep.\end{tabular}} &
    \textbf{\begin{tabular}[c]{@{}l@{}}Locally\\ Con.\end{tabular}} &
    \textbf{\begin{tabular}[c]{@{}l@{}}Feature\\ Dep.\end{tabular}} &
    \textbf{Uniform} &
    \textbf{\begin{tabular}[c]{@{}l@{}}Class \\ Dep.\end{tabular}} &
    \textbf{\begin{tabular}[c]{@{}l@{}}Locally \\ Con.\end{tabular}} &
    \textbf{\begin{tabular}[c]{@{}l@{}}Feature \\ Dep.\end{tabular}} \\ \hline
  \textbf{5} &
    85\% &
    84\% &
    84\% &
    \cellcolor[HTML]{67FD9A}91\% &
    88\% &
    87\% &
    87\% &
    \cellcolor[HTML]{67FD9A}93\% &
    91\% &
    91\% &
    90\% &
    \cellcolor[HTML]{F56B00}89\% \\ \hline
  \textbf{15} &
    75\% &
    75\% &
    73\% &
    \cellcolor[HTML]{67FD9A}93\% &
    77\% &
    78\% &
    77\% &
    \cellcolor[HTML]{67FD9A}95\% &
    90\% &
    90\% &
    90\% &
    \cellcolor[HTML]{F56B00}84\% \\ \hline
  \textbf{25} &
    66\% &
    66\% &
    65\% &
    \cellcolor[HTML]{67FD9A}89\% &
    68\% &
    68\% &
    69\% &
    \cellcolor[HTML]{67FD9A}91\% &
    90\% &
    89\% &
    85\% &
    \cellcolor[HTML]{F56B00}80\% \\ \hline
  \textbf{35} &
    57\% &
    57\% &
    56\% &
    \cellcolor[HTML]{67FD9A}83\% &
    58\% &
    60\% &
    60\% &
    \cellcolor[HTML]{67FD9A}83\% &
    89\% &
    89\% &
    83\% &
    \cellcolor[HTML]{F56B00}76\% \\ \hline
  \textbf{45} &
    48\% &
    49\% &
    48\% &
    \cellcolor[HTML]{67FD9A}78\% &
    48\% &
    49\% &
    54\% &
    \cellcolor[HTML]{67FD9A}80\% &
    88\% &
    86\% &
    70\% &
    \cellcolor[HTML]{F56B00}67\% \\ \hline
  \textbf{55} &
    39\% &
    40\% &
    40\% &
    \cellcolor[HTML]{67FD9A}73\% &
    39\% &
    40\% &
    44\% &
    \cellcolor[HTML]{67FD9A}76\% &
    87\% &
    80\% &
    68\% &
    \cellcolor[HTML]{F56B00}57\% \\ \hline
  \textbf{65} &
    30\% &
    35\% &
    37\% &
    \cellcolor[HTML]{67FD9A}71\% &
    30\% &
    37\% &
    40\% &
    \cellcolor[HTML]{67FD9A}74\% &
    85\% &
    \cellcolor[HTML]{F56B00}47\% &
    50\% &
    48\% \\ \hline
  \textbf{75} &
    21\% &
    36\% &
    31\% &
    \cellcolor[HTML]{67FD9A}61\% &
    22\% &
    36\% &
    35\% &
    \cellcolor[HTML]{67FD9A}65\% &
    81\% &
    \cellcolor[HTML]{F56B00}21\% &
    35\% &
    43\% \\ \hline
  \end{tabular}
  \caption{Train, validation and test accuracies for different noise ratios for MNIST-Fashion. High accuracies for train and validation sets are marked with green and low accuracies for test set are marked with red.}
\end{table*} \label{tbl:noiseratios}

\subsection{Dataset Size}
We experimented with different dataset sizes in order to investigate if the number of samples affects the performance in the presence of various label noise types. 1\%, 10\%, 40\%, 70\%, and 100\% of data from the dataset is taken, and the network is trained on this subset. As can be seen in \autoref{tbl:datasize}, increasing the size of the dataset results in an increase of accuracy in all noise types besides feature-dependent noise. This is due to the correlation of noise to the real feature distribution of data. With the increasing number of data points, the negative effect of uncorrelated or \textit{less correlated} noise types is diminishing, since updates caused by noisy samples are overwhelmed by gradient updates from clean samples.

\subsection{Hyper-Parameters}
We investigated the effect of dropout layer and model complexity for different noise types, as presented below.
\begin{itemize}
  \item \textbf{Dropout:} We tried out performances of the same architectures with and without dropout layers. Increase in performances are given in \autoref{tbl:dropout}. As can be seen, dropout boosts the performance of feature independent noises since they are randomly distributed in the feature domain. However, for structured noises, such as locally-concentrated and feature-dependent, dropout doesn't provide a significant boost in performance.
  \item \textbf{Network Complexity:} In order to see if the impact of noise depends on the model architecture, we repeated all experiments on the MNIST-Fashion dataset with three-layer MLP architecture. Obtained results are observed to be consistent with CNN architecture results. Therefore, the presented results can be evaluated as model-agnostic.
\end{itemize}

\begin{table}[]
  \centering
  \begin{tabular}{|l|llll|}
  \hline
  \multicolumn{1}{|c|}{\textbf{\begin{tabular}[c]{@{}c@{}}Noise\\ Rates\\ (\%)\end{tabular}}} &
    \multicolumn{1}{c}{\textbf{Uniform}} &
    \multicolumn{1}{c}{\textbf{\begin{tabular}[c]{@{}c@{}}Class\\ Dep.\end{tabular}}} &
    \multicolumn{1}{c}{\textbf{\begin{tabular}[c]{@{}c@{}}Locally\\ Con.\end{tabular}}} &
    \multicolumn{1}{c|}{\textbf{\begin{tabular}[c]{@{}c@{}}Feature\\ Dep.\end{tabular}}} \\ \hline
  25 & 18\% & 21\% & 1\%   & 6\% \\ \hline
  35 & 80\% & 18\% & 6\% & 2\% \\ \hline
  45 & 21\% & 30\% & 1\%   & 6\% \\ \hline
  55 & 39\% & 36\% & 17\% & 5\% \\ \hline
  \end{tabular}
  \caption{Increase ratio in test accuracies due to usage of dropout layer for CIFAR100 dataset for different types of noises at different noise rates.}
  \end{table} \label{tbl:dropout}

\subsection{Noise-Robust Algorithms}
We tested noise robust algorithms from literature on given label noise types to see their effectiveness. Used methods can be listed as below:

\begin{itemize} 
  \item \textbf{Forward Loss Correction} \cite{patrini2017making}: Proposes to use confusion matrix for loss correction. In their work, authors propose methods to estimate the true confusion matrix. However, in our case, since we already know the true confusion matrix of synthetic noise, it is directly used without approximation.
  \item \textbf{Bootstrap} \cite{reed2014training}: Checks the consistency among network predictions and given labels to test its noisiness.
  \item \textbf{D2L} \cite{ma2018dimensionality}: Monitors complexity of network to see if it starts to overfit noise and uses this information to regularize learning.
  \item \textbf{Co-Teaching} \cite{han2018co}: Uses two networks in pair. Each network back-propagate on instances where its pair has a small loss.
\end{itemize}

Experimental results at \autoref{tbl:noiserobust} shows that each algorithm behaves differently for different label noise type. For example, forward loss and D2L gives the best performance on uniform noise while co-teaching is the best for option other types of noise. Interestingly, the D2L method gives one of the best performances at uniform noise but achieves worse than normal training for locally-concentrated and feature-dependent noises. Overall, we found that co-teaching is the most efficient method among the listed methods. 
\begin{table}[]
  \centering
  \begin{tabular}{|c|ccccc|}
  \hline
  \multicolumn{1}{|l|}{} &
    \multicolumn{1}{l}{\textit{\textbf{Normal}}} &
    \multicolumn{1}{l}{\textit{\textbf{Forward}}} &
    \multicolumn{1}{l}{\textit{\textbf{Bootstrap}}} &
    \multicolumn{1}{l}{\textit{\textbf{D2L}}} &
    \multicolumn{1}{l|}{\textit{\textbf{Co-Teach.}}} \\ \hline
  \textbf{Uniform}                                              & 52\% & 59\% & 57\% & 59\% & 57\% \\ \hline
  \textbf{\begin{tabular}[c]{@{}c@{}}Class Dep.\end{tabular}}   & 52\% & 54\% & 51\% & 54\% & 57\% \\ \hline
  \textbf{\begin{tabular}[c]{@{}c@{}}Locally Con.\end{tabular}} & 31\% & 35\% & 31\% & 26\% & 57\% \\ \hline
  \textbf{\begin{tabular}[c]{@{}c@{}}Feature Dep.\end{tabular}} & 38\% & 41\% & 39\% & 28\% & 55\% \\ \hline
  \end{tabular}
  \caption{Test accuracies of noise-robust algorithms on CIFAR100 dataset for different types of noisy labels for noise ratio of 35\%. Normal column represents training results without additional noise-robust algorithm applied.}
\end{table} \label{tbl:noiserobust}


	\section{Conclusion} \label{conclusion}
In this work, we analyzed the negative impacts of label noise on deep learning algorithms, depending on the noise type. Even though lots of works in literature are devoted to overcoming the label noise phenomenon, it is problematic that each method is tested under their own generated synthetic label noise. Therefore, we propose a generic label corruption algorithm with synthetic feature-dependent label noise. The proposed methodology uses \textit{distillation} technique to create a sparse distribution of data in the learned feature domain and therefore emphasizes similarities among data samples. We investigated the negative effects of each type of label noise form a different perspective. Interestingly, it is observed that feature dependent noise shows similar behavior in the training and validation phase while resulting in much lower test accuracy. Therefore, it is much harder to evaluate the progress of the network in case of feature-dependent noise by just checking outputs on noisy training and validation sets. Moreover, it is seen that noise-robust algorithms behave differently for different types of noises. Therefore, it is important to consider the noise type while evaluating the proposed methodology. For future researches, we provided generated label noise at different ratios for common benchmark datasets: MNIST, MNIST-Fashion, CIFAR10, CIFAR100. However, the proposed algorithm can be used to generate feature-dependent noise on any given dataset.

	\bibliographystyle{IEEEtran}
	\bibliography{paperrefs} 
	
\end{document}